\documentclass[lettersize,journal]{IEEEtran}

\usepackage{cite}
\usepackage{amsmath,amssymb,amsfonts}
\usepackage{algorithmic}
\usepackage{graphicx}
\usepackage{makecell}
\usepackage{multirow}
\usepackage{color}
\usepackage{algorithm,algorithmic}

\usepackage{textcomp}
\usepackage{wrapfig}
\usepackage[colorlinks,
            linkcolor=red,
            anchorcolor=blue,
            citecolor=green 
            ]{hyperref}

\begin{document}

\title{Ivan-ISTD: Rethinking Cross-domain Heteroscedastic Noise Perturbations in Infrared Small Target Detection }

\author{Yuehui Li, Yahao Lu, Haoyuan Wu, Sen Zhang, Liang Lin,~\IEEEmembership{Fellow,~IEEE}, Yukai Shi
\thanks{Y. Li, Y. Lu, H. Wu and Y. Shi are with School of Information Engineering, Guangdong University of Technology, Guangzhou, 510006, China (email: liyuehui77161@gmail.com; 2112303120@mail2.gdut.edu.cn; bridgesness@gmail.com; ykshi@gdut.edu.cn)

S. Zhang is with TikTok, ByteDance Inc, Sydney, NSW 2000, Australia (email: senzhang.thu10@gmail.com).

L. Lin is with School of Data and Computer Science, Sun Yat-sen University, Guangzhou, 510006, China (email: linliang@ieee.org).

}

}



\maketitle

\begin{abstract}

In the multimedia domain, Infrared Small Target Detection (ISTD) plays a important role in drone-based multi-modality sensing. To address the dual challenges of cross-domain shift and heteroscedastic noise perturbations in ISTD, we propose a doubly wavelet-guided Invariance learning framework(Ivan-ISTD). In the first stage, we generate training samples aligned with the target domain using Wavelet-guided Cross-domain Synthesis. This wavelet-guided alignment machine accurately separates the target background through multi-frequency wavelet filtering. In the second stage, we introduce Real-domain Noise Invariance Learning, which extracts real noise characteristics from the target domain to build a dynamic noise library. The model learns noise invariance through self-supervised loss, thereby overcoming the limitations of distribution bias in traditional artificial noise modeling. Finally, we create the Dynamic-ISTD Benchmark, a cross-domain dynamic degradation dataset that simulates the distribution shifts encountered in real-world applications. Additionally, we validate the versatility of our method using other real-world datasets. Experimental results demonstrate that our approach outperforms existing state-of-the-art methods in terms of many quantitative metrics. In particular, Ivan-ISTD demonstrates excellent robustness in cross-domain scenarios. The code for this work can be found at: \url{https://github.com/nanjin1/Ivan-ISTD}.
\end{abstract}


\section{Introduction}
\begin{figure}[t]
\centering
\includegraphics[width=0.95\columnwidth]{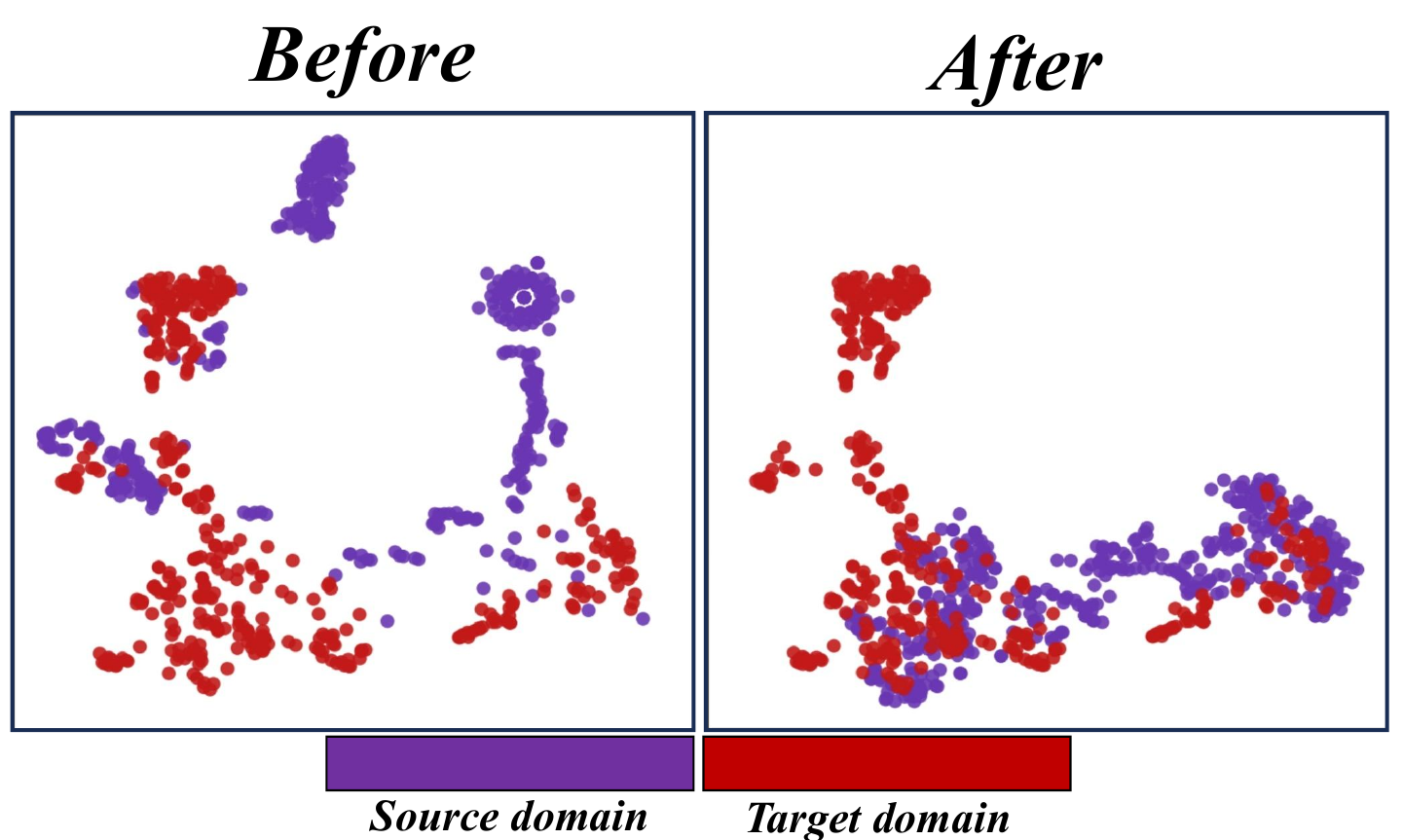}
\caption{A significant distribution shift between the source and target domains. Our transfer background method helps mitigate this domain shift issue.}
\label{fig1}
\vspace{-3mm}
\end{figure}

\begin{figure}[t]
\centering
\includegraphics[width=0.99\columnwidth]{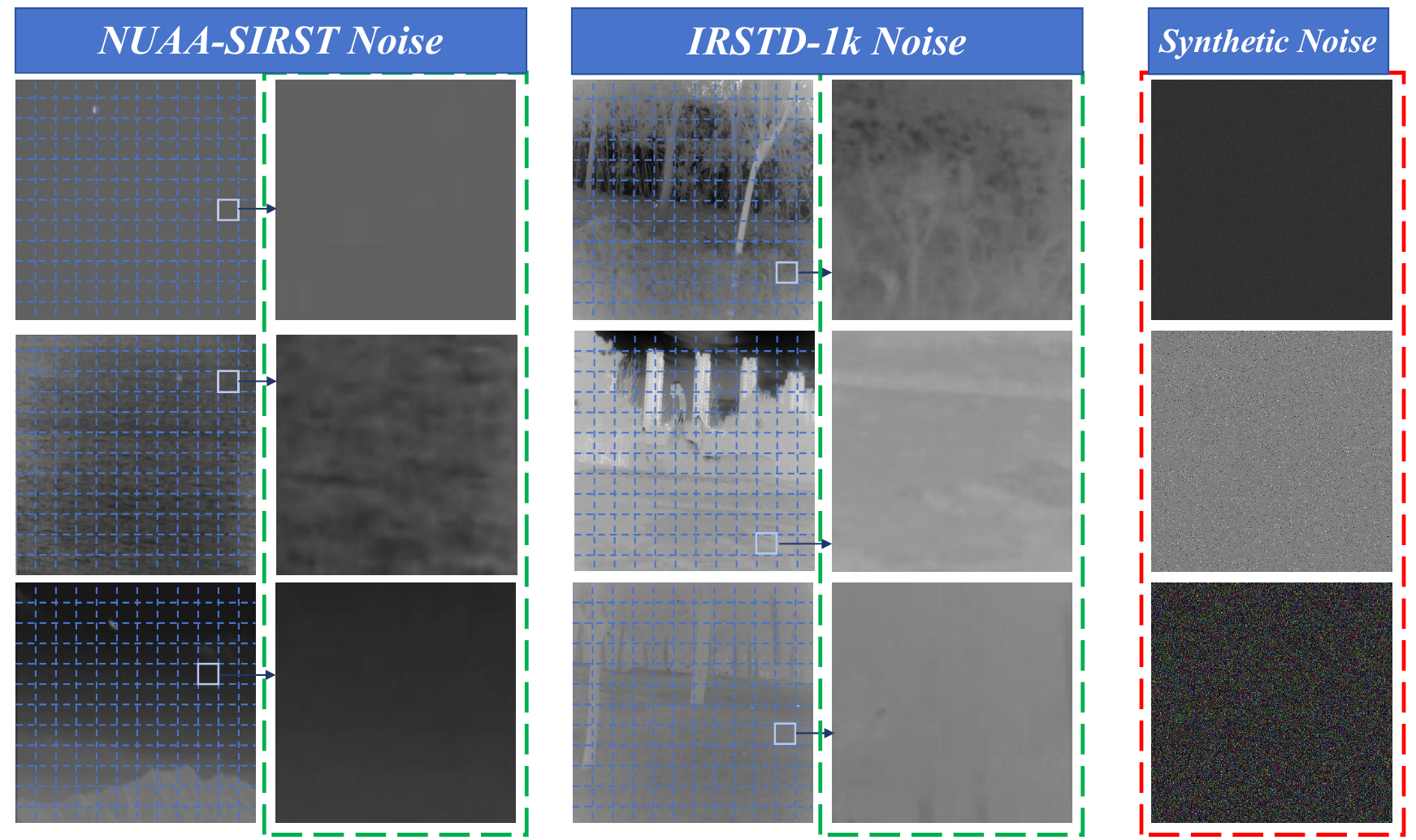}
\caption{Real domain noise (green) and artificially synthesized noise (red). The real domain noise exhibits greater variability, whereas the synthetic noise is uniform and cannot capture the dynamic heteroscedastic nature of noise in real-world situations.}
\label{fig2}
\vspace{-3mm}
\end{figure}

\IEEEPARstart{I}{n} the multimedia domain, Infrared Small Target Detection (ISTD) plays a crucial role in drone-based multi-modality sensing and applications~\cite{luo2025spatial,lin2024learning,chen2023glrt,weng2025dronesr}. However, achieving consistent and robust detection performance in complex target environments remains an ongoing challenge~\cite{zhao2022single}. Existing deep learning models are often impacted by real-world data distribution shifts, including:
\begin{itemize}
    \item[$\bullet$] \textbf{Background-induced Domain Shift}: As illustrated in Fig.~\ref{fig1}, differences in background environments~\cite{wang2019miss,chen2023glrt} cause a substantial distribution shift between the source and target domains~\cite{federici2021information}. As a result, the features learned by the model during training are often not transferable to target domain data with different backgrounds~\cite{wang2025noise}.
    \item[$\bullet$] \textbf{Cross-domain Heteroscedastic Noise Perturbations}: As shown in Fig.~\ref{fig2}, noise characteristics, such as type, intensity, and distribution, can vary significantly between source and target domains due to differences in equipment and environmental factors~\cite{song2022learning,shi2024nitedr,le2005heteroscedastic,shi2025crossfuse}. Traditional methods typically rely on artificially designed uniform noise augmentation~\cite{chen2013local,meng2023robust}, but they fall short in capturing the dynamic and heteroscedastic nature of real-world noise~\cite{pal2023understanding}.
\end{itemize}

\begin{figure*}[t]
\centering
\includegraphics[width=0.98\linewidth]{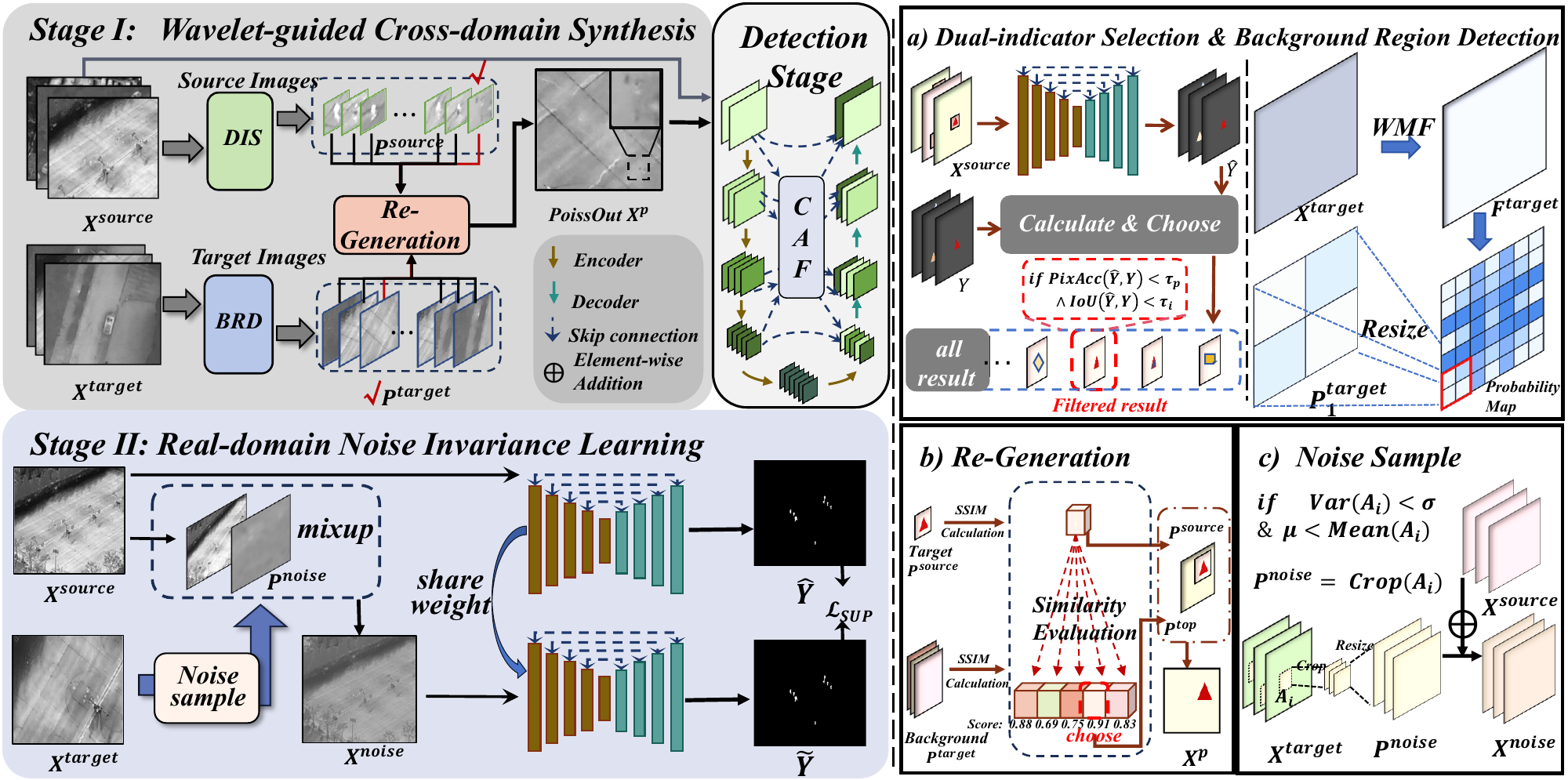}
\caption{The overall framework of our method. Stage I presents the complete Wavelet-guided Cross-domain Synthesis framework, consisting of (a) and (b). (a) illustrates the process of extracting high-value small targets from the source set (right image) and unsupervised background sample extraction (left image). (b) describes the synthesis of new samples using SSIM ranking and Poisson blending techniques. Stage II outlines the process of Real-domain Noise Invariance Learning. It involves unsupervised real noise sampling from target images, minimizing loss to enhance the model's robustness to cross-domain noise. (c) shows the detailed process of extracting noise from the target set. }
\label{fig4}
\vspace{-3mm}
\end{figure*}

These challenges result in a performance gap between source and target domain data. Traditional techniques, such as synthetic noise augmentation~\cite{nishi2021augmentation,lu2024sirst} and static data preprocessing~\cite{tsoi1995static,xi2022combined}, are ineffective in addressing the complex noise distributions and dynamic background changes encountered in real-world testing conditions.

In recent years, researchers have been exploring methods to build hierarchical feature representations and facilitate cross-layer feature reuse. For instance, in the multimedia domain, HCF-Net~\cite{hcf} addresses multi-scale background interference by using a hierarchical context fusion strategy, Lin~\cite{lin2023learning} introduces a large-kernel encoder and a shape-guided decoder through learning shape-biased representations, which incorporate shape information to improve the localization of faint targets, though this method depends on supervised training; Luo~\cite{luo2025spatial} presents a spatio-temporal aware unsupervised network that boosts adaptability to dynamic scenes; and SCTransNet~\cite{sctransnet} combines spatial attention with channel recalibration mechanisms. However, current research is limited by its reliance on fixed training datasets, which hinders adaptability to unknown target domains.

To address domain shifts arising from background environmental differences and cross-domain heteroscedastic noise, the CORAL~\cite{zhang2023coral} feature alignment method, based on second-order statistics, reduces domain differences by aligning the statistical distributions of source and target domains. On the other hand, the RandConv~\cite{xu2020robust,choi2023progressive} increases the diversity of training data through artificially generated noise or style perturbations. However, both methods are constrained by the assumption of static noise, which fails to capture the dynamic, complex noise distributions encountered in real-world environments~\cite{liao2024denoising,han2020towards}.

To overcome these limitations, we propose a Doubly Wavelet-guided Invariance Learning Framework. In the first stage, we introduce Wavelet-guided~\cite{mallat2002theory} Cross-domain Synthesis, a strategy that enables cross-domain adaptation during training, ensuring stable small target detection without the need for additional inference adjustments. In the second stage, Real-domain Noise Invariance Learning extracts real noise features from the target domain to build a dynamic noise library. By mixing noise data and applying self-supervised loss constraints, the model learns to be invariant to noise.

Additionally, we have developed a cross-domain dynamic degradation dataset for small targets in UAV infrared imaging:  Dynamic-ISTD. This dataset contains training and testing sets from various domains, designed to simulate the distribution shifts typically encountered in real-world scenarios.

In summary, the key contributions of this paper are as follows:

\begin{itemize}
    
    \item[$\bullet$] We propose the Wavelet-guided Cross-domain Synthesis strategy, which allows the model to adapt to target domain features at the data space level during training, thus enhancing cross-domain generalization.
    \item[$\bullet$] We present the Real-domain Noise Invariance Learning strategy, which enables the model to adapt to the noise characteristics of the target domain at the feature space level.
    \item[$\bullet$] We introduce a new cross-domain dynamic degradation dataset, Dynamic-ISTD Benchmark, specifically for UAV infrared small targets. The dataset includes training and testing sets from different domains to replicate the cross-domain distribution shifts that are likely to occur in practical applications.
    \item[$\bullet$] Through cross-domain validation experiments, we show the effectiveness and broad applicability of our method across various datasets.
\end{itemize}

\begin{figure*}[ht]
\centering
\includegraphics[width=0.99\linewidth]{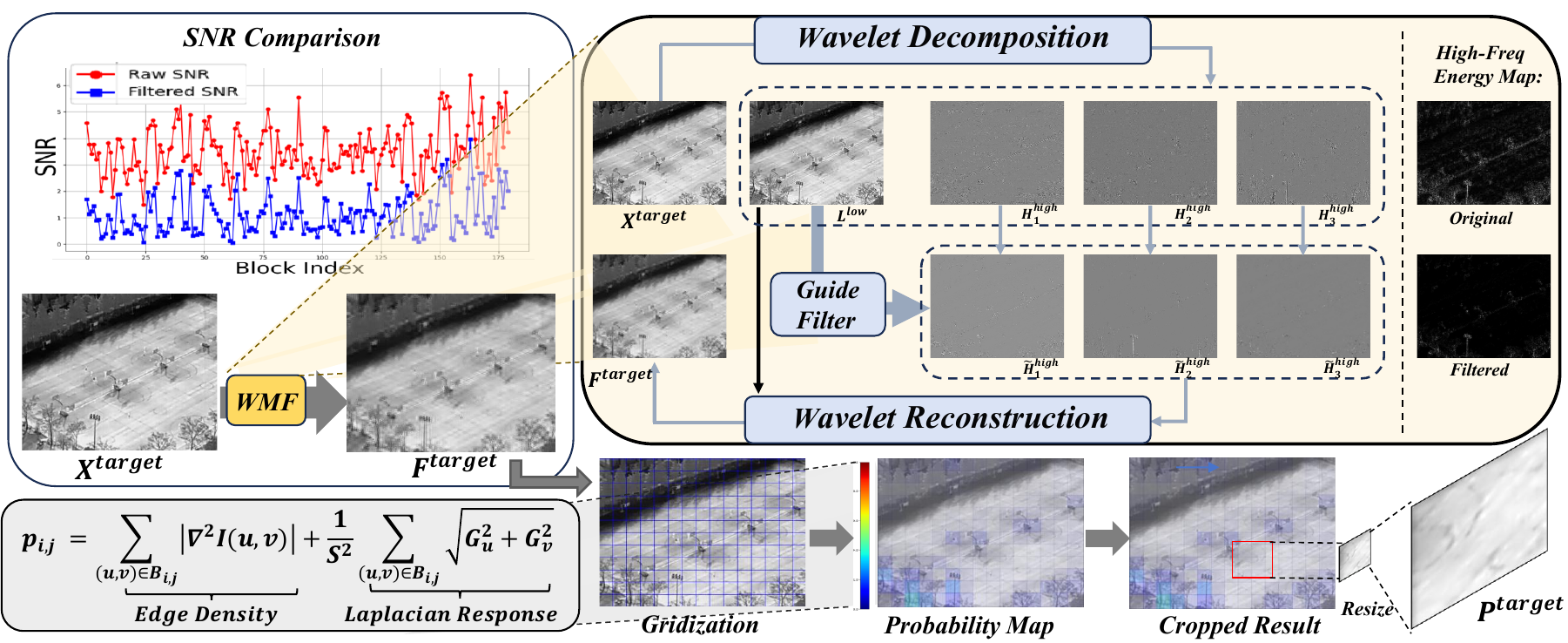}
\caption{(a) illustrates the proposed unsupervised Background Region Detection framework. After applying Wavelet Multi-frequency Filtering (WMF) and calculating small target probabilities based on edge density and Laplacian operator, we select appropriate background samples. (b) shows the detailed implementation of WMF, where $L^{low}$ is the low-frequency sub-band obtained through wavelet decomposition, and $\{H_k^{high}\}^3_{k=1}$ is the high-frequency sub-band.}
\label{fig5}
\vspace{-3mm}
\end{figure*}
\section{Related work}
\subsection{Single-frame Infrared Small Target Detection}
The precise detection and segmentation of small, weak targets in complex infrared backgrounds remain a major challenge. The inherent limitations of infrared imaging cause non-uniform background interference, significantly degrading the performance of traditional detection algorithms across different environments. Traditional methods, primarily focused on local contrast analysis (such as Tophat~\cite{bai2010analysis,deng2021infrared} and IPI~\cite{gao2013infrared,xi2022combined}), struggle to handle dynamic background textures due to their inability to model multi-scale features effectively. Deep learning approaches, which build hierarchical feature representations, have shown considerable advantages in target representation learning\cite{luo2025spatial}. DNA-Net\cite{
DNA} improves small target responses by reusing features across layers, HCF-Net\cite{hcf} uses a hierarchical context fusion strategy to address multi-scale background interference\cite{lin2024learning}, and SCTransNet~\cite{sctransnet} innovatively combines spatial attention with channel recalibration mechanisms, maintaining robust detection even in the presence of complex thermal noise. Despite these advancements, the lack of real-world environment modeling and the sensitivity to domain shifts continue to limit their practical use~\cite{zhao2022single}.

\subsection{Domain Shift Challenges}
The main challenge of domain shift lies in the complex and dynamic differences in data distributions between the source and target domains~\cite{gao2024appearance}, such as changes in lighting, imaging conditions, or sensor noise~\cite{buades2005review,baek2009integrated}. These differences often lead to a significant drop in model performance when applied to real-world scenarios.
In recent years, researchers have introduced various strategies to counteract the performance degradation caused by domain shifts, including adaptive feature alignment~\cite{zhang2023enhanced,guan2021domain,pan2009survey}. CORAL~\cite{sun2016return,zhang2023coral} uses second-order statistics, minimizes feature distribution differences by aligning the covariance matrices of the source and target domains. However, these methods typically rely on explicit domain label information to differentiate between various imaging conditions, making them difficult to apply in practical scenarios where the target domain lacks or has limited labels~\cite{fang2024source,zhang2021survey}.

\section{Methodology}
\label{sec:guidelines} 
As shown in Fig.~\ref{fig4}, In this section, we introduces a new two-stage optimization framework. In the first stage, the framework aligns domains in the data space. In the second stage, it reduces domain differences and noise sensitivity in the feature space. 

\subsection{Wavelet-guided Cross-domain Sample Synthesis}
As shown in Fig.~\ref{fig4}, we use Background Region Detection (BRD) and Dual-indicator Selection (DIS) to carry out unsupervised background filtering and extract high-value targets.

\textbf{Background Region Detection(BRD)}. We use Wavelet Multi-frequency Filtering (WMF) to split the target images into sub-bands at different frequencies. The low-frequency baseband $L^{low}$ captures the primary structural details of the image, while the high-frequency bands $\{H_k^{high}\}^3_{k=1}$ capture edge details and noise. By using the low-frequency image to guide the high-frequency sub-bands, we perform edge-aware filtering that suppresses noise while preserving the edge details accurately:
\begin{equation}
\tilde{H}_k^{high}  = H_k^{high} \cdot \frac{|\nabla L^{low}|}{\max(|\nabla L^{low}|) + \epsilon},
\end{equation}
where $H_k^{high}$ and $\tilde H_k^{high}$ represent the high-frequency detail bands before and after filtering, $L^{low}$ is the low-frequency baseband, and $\epsilon$ is a regularization term to prevent numerical overflow. Additionally, we split the $F^{target}$ into multiple non-overlapping subblocks $B_{i,j}$. For each subblock, we calculate its edge density and Laplacian operator. The features are then normalized to the range $[0,1]$ and combined to yield the probability $p_{i,j}$ of the background tendency region.
\begin{equation}
p_{i,j} =  \sum_{(u,v) \in B_{i,j}} \left| \nabla^2 I(u,v) \right| + \frac{1}{S^2} \sum_{(u,v) \in B_{i,j}} \sqrt{G_u^2 + G_v^2},
\end{equation}
where $(i,j)$ represents the subblock index, $(u,v)$ is the pixel index within the image, $S$ is the subblock size, and $G_u$ and $G_v$ are the horizontal and vertical gradient fields of the subblock, respectively. $\nabla^2$ is the Laplacian operator. Low-probability regions correspond to areas with low-texture background characteristics. A preseted threshold $\tau_b$ is applied to select background subblocks $P_{}^{target} \in R^{126\times126}$ from the original image $X_{}^{target}$. These background regions are then upsampled to the original resolution using bilinear interpolation, ensuring geometric consistency with the input image. As shown in Fig.~\ref{fig5}, the image after WMF exhibits a higher signal-to-noise ratio, which suggests that WMF effectively distinguishes real edges from random noise in the high-frequency sub-bands. And the workflow can be found in Algorithm~\ref{alg:code1}.

\begin{figure}[t]
\centering
\includegraphics[width=0.99\columnwidth]{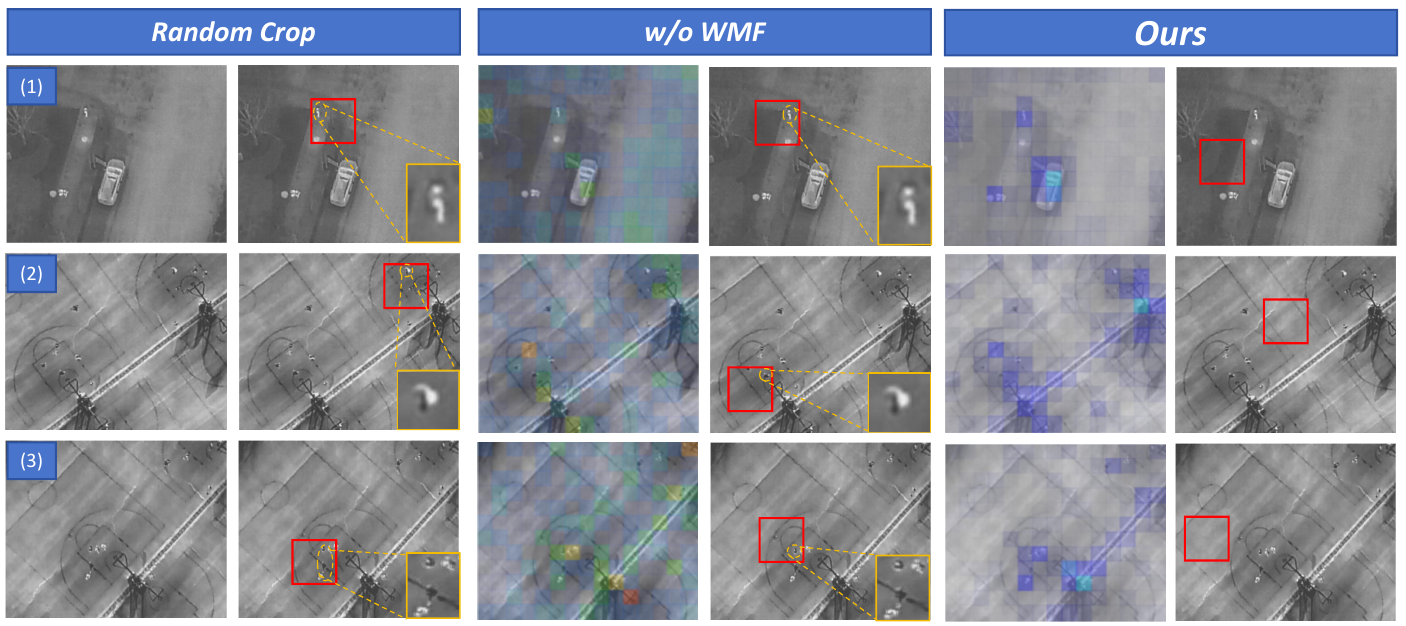}
\caption{Visualization results of Background Region Detection methods on target data. The left column displays the background region tendency map, while the right column presents the cropped region. The red boxes indicate the cropping results from different methods, and the yellow dashed circles highlight cases where other methods misclassified the background, including unnecessary small targets within the background. The yellow boxes are used to clearly emphasize the small targets.}
\label{fig6}
\vspace{-3mm}
\end{figure}

As shown in Fig.~\ref{fig6}, we use low-frequency structure guidance to filter high-frequency sub-bands. It effectively reduces random noise interference. It also helps to distinguish small targets from subtle background textures. This provides reliable prior information for cross-domain data optimization.

In addition, we use a dual-indicator selection strategy to build the difficult target set. During each iteration, the model parameter $f_\theta$ generates a predicted mask $\hat Y=f_\theta( X^{\text{source}})$ for the training set $X^{\text{source}}$ . We then evaluate the candidate regions using pixel accumulation (PixAcc) and image quality (IoU). The final difficult target set $P^{source}$ is selected based on these evaluations.

\begin{equation}
P^{\text{source}} =
\begin{cases} 
\text {Select} & \text{if } \text{PixAcc}(\hat Y,Y) < \tau_p \wedge \text{IoU}(\hat Y,Y) < \tau_i
\\
\varnothing & \text{otherwise}
\end{cases}
\end{equation}
Here, $\tau_p$ and $\tau_i$ represent the threshold values for PixAcc and IoU, respectively.

\begin{figure}[t]
\centering
\includegraphics[width=0.99\columnwidth]{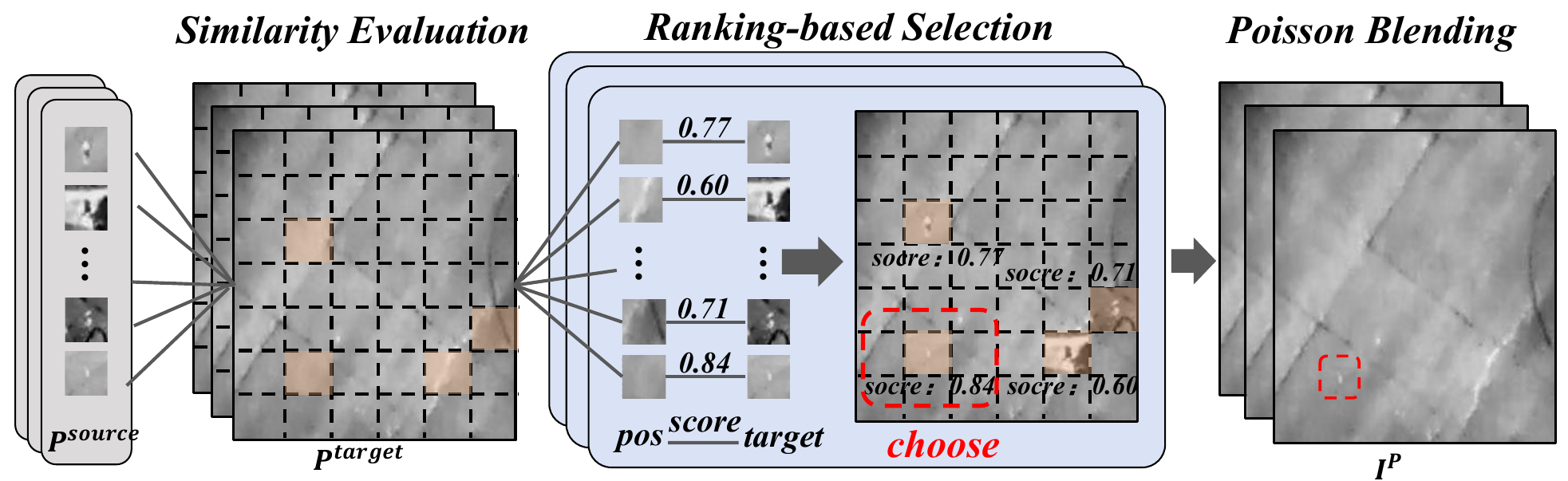}
\caption{Illustration of Structural Similarity (SSIM) evaluation and fusion. First, the color and texture similarities between the target images and source image blocks are calculated to generate SSIM scores for the candidate regions. The regions are then ranked by their scores, and the best matching region is selected for target embedding.}
\label{fig7}
\vspace{-3mm}
\end{figure}

\textbf{Re-Generation.} As shown in Fig.~\ref{fig7}, we grid-segment the candidate background region 
$P_{\text{}}^{\text{target}}$ to generate a set of local windows $A^{target}$. We then calculate the structural similarity (SSIM) between these windows and the candidate small target set $P_{\text{}}^{\text{source}}$. The regions are sorted by SSIM values in descending order, and the best matching region ${A}^{top}$ is selected.

\begin{equation} 
\mathbf{A}^{top}= {TOP_{SSIM}}\mathbf{(A_{\text{}}^{\text{target}}}, \mathbf{P_{\text{}}^{\text{source}}}).
\end{equation}
where $A^{target}$ refers to the local windows in the candidate background, and $P_{\text{}}^{\text{source}}$ is the set of candidate targets. The workflow can be found in Algorithm~\ref{alg:code2}. To remove color discrepancies and seam artifacts, we apply Poisson Fusion to optimize the gradient continuity.

\begin{equation}
\min_{{X}^{p}} \iint_{A^{top}} \left| \nabla  {{X}^{p}} - \nabla{P^{\text{source}}} \right|^2 \, du \, dv
, {I}^{a}|_{\partial A^{top}} = {I}^{a*}|_{\partial A^{top}}.
\end{equation}
Here,${X}^{p}$ refers to the synthesized image, and $A^{top}$ is the target coverage area. The pixel mappings ${I}^{a}$ and ${I}^{a*}$ inside and outside the synthesized image are kept consistent at the boundary $\partial A^{top}$.

\begin{algorithm}[!ht]
    \renewcommand{\algorithmicrequire}{\textbf{Input:}}
    \renewcommand{\algorithmicensure}{\textbf{Output:}}
    \caption{Background Region Detection}
    \label{alg:code1}
    \begin{algorithmic}
        \REQUIRE $X^{\text{target}} \in \mathbb{R}^{640 \times 512}$: Grayscale image
        \ENSURE $P^{\text{target}} \in \mathbb{R}^{126 \times 126}$: Cropped low-probability region
        
        \STATE $\triangleright$ Wavelet Multi-frequency Filtering
        \STATE $[L^{low}, \{H_k^{high}\}_{k=1}^3] = \text{WaveletDecompose}(X^{\text{target}}, \psi, L)$
        \FOR{$k \gets 1$ to $3$}
            \STATE $\tilde{H}_k^{high} = H_k^{high} \cdot \frac{|\nabla L^{low}|}{\max(|\nabla L^{low}|) + \epsilon}$
        \ENDFOR
        \STATE $F^{\text{target}} = \text{WaveletReconstruct}(L^{low}, \{\tilde{H}_k^{high}\}_{k=1}^3, \psi)$
        
        \STATE $\triangleright$ Block Feature Extraction
        \STATE Divide $F^{\text{target}}$ into $(g_h, g_w)$ grid blocks
        \FOR{each block $B_{i,j}$}
            \STATE $p_{i,j} = 0.5 \cdot \text{EdgeDensity}(B_{i,j}) + 0.5 \cdot \text{LaplacianResponse}(B_{i,j})$
        \ENDFOR
        
        \STATE $\triangleright$ Optimal Low-Probability Region Detection
        \STATE $P^{\text{target}} = \text{CropRegion}(\{B_{i,j} \mid p_{i,j} < \tau\})$
    \end{algorithmic}
\end{algorithm}

\begin{algorithm}[!ht]
    \renewcommand{\algorithmicrequire}{\textbf{Input:}}
    \renewcommand{\algorithmicensure}{\textbf{Output:}}
    \caption{Re-Generation Algorithm}
    \label{alg:code2}
    \begin{algorithmic}
        \REQUIRE $N$ backgrounds $P_N^{\text{target}}$, $M$ targets $P_M^{\text{source}}$, SSIM threshold $t$
        \ENSURE $M$ composite images $I^P$
        
        \STATE $U_m = 0$ for each target
        \FOR{$n \gets 0$ to $N$}
            \FOR{$m \gets 0$ to $M$}
                \STATE $(Pos_{mn}, Score_{mn}) = \text{SSIM}(P_m^{\text{source}}, P_n^{\text{target}})$
            \ENDFOR
            \STATE $Candidates = [(m, Pos, Score) \mid Score \geq t]$ sorted by Score
            \FOR{each $(m, Pos, Score)$ in $Candidates[0:10]$ (shuffled)}
                \IF{$U_m < MaxUsage$}
                    \STATE $I^P = \text{PoissonBlend}(P_m^{\text{source}}, P_n^{\text{target}}, Pos)$
                    \STATE $U_m = U_m + 1$
                    \STATE \textbf{break}
                \ENDIF
            \ENDFOR
        \ENDFOR
    \end{algorithmic}
\end{algorithm}

\begin{figure*}[!t]
\centering
\includegraphics[width=0.95\linewidth]{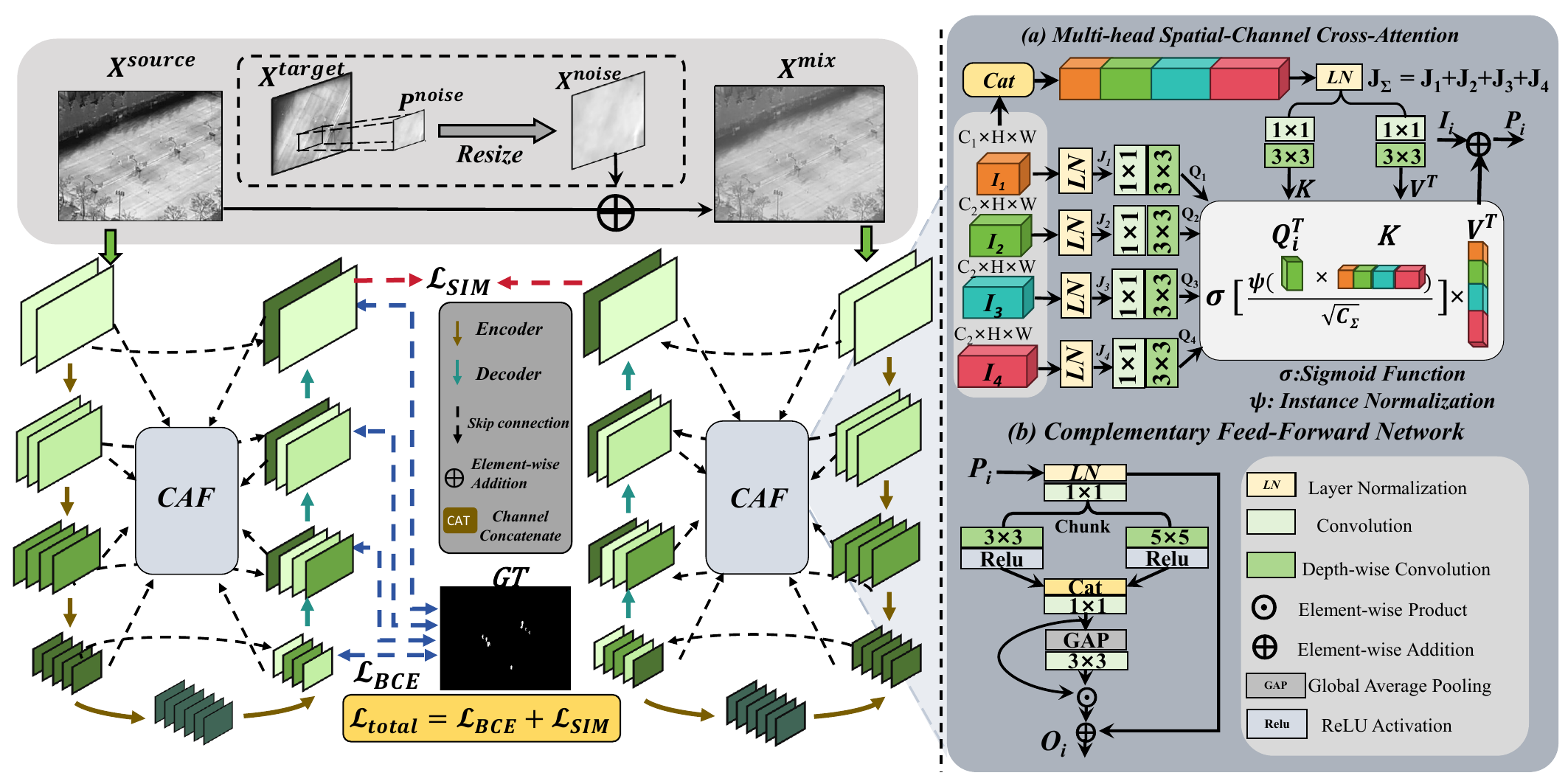}
\caption{The overall process of Real-domain Noise Invariance Learning. The upper-left section depicts the Noise Mixing process, where appropriate noise is extracted from the target set and mixed into the source set to create a new dataset.The lower-left section shows the dual-branch network architecture with shared weights, optimized through a combination of supervised and self-supervised losses to learn denoising features. The images on the right (a) and (b) show the modules within CAF.}
\label{fig8}
\vspace{-3mm}
\end{figure*}
\subsection{Real-world Domain Noise Invariance Learning }
\textbf{Network Structure.} A five-layer residual downsampling module $\{E_i\}^5_{i=1}$ is used in the encoder to extract multi-scale high-level features, where $E_i \in \mathbb{R}^{H/2^i \times W/2^i\times C_i}$, and low-dimensional semantic representations are generated through block embedding. In the decoder, spatial resolution is progressively restored using upsampling $\{D_i\}^4_{i=1}$ and skip connections. Binary cross-entropy loss is applied to the output of each decoding layer. Multi-scale feature fusion is employed to aggregate contextual information from different levels, generating the final prediction. The total loss is computed through weighted summation. Specifically, the supervised loss $L_{BCE}$ is composed of losses from all scales:
\begin{equation}
\mathcal{L}_{\text BCE} = \sum_{i=1}^{4} \lambda_i \cdot {BCE(Y_i, \hat{Y}_i)}.
\end{equation}

Where $BEC(\cdot)$ refers to the binary cross-entropy loss function. $Y_i$ is the true label for the $i$-th scale, while 
$\hat{Y}_i$ is the predicted output for that scale. $L_{\text{BCE}}$ represents the supervised loss, and $\lambda_i$ is the weight coefficient for each scale.

\begin{figure*}[!t]
\centering
\includegraphics[width=0.95\linewidth]{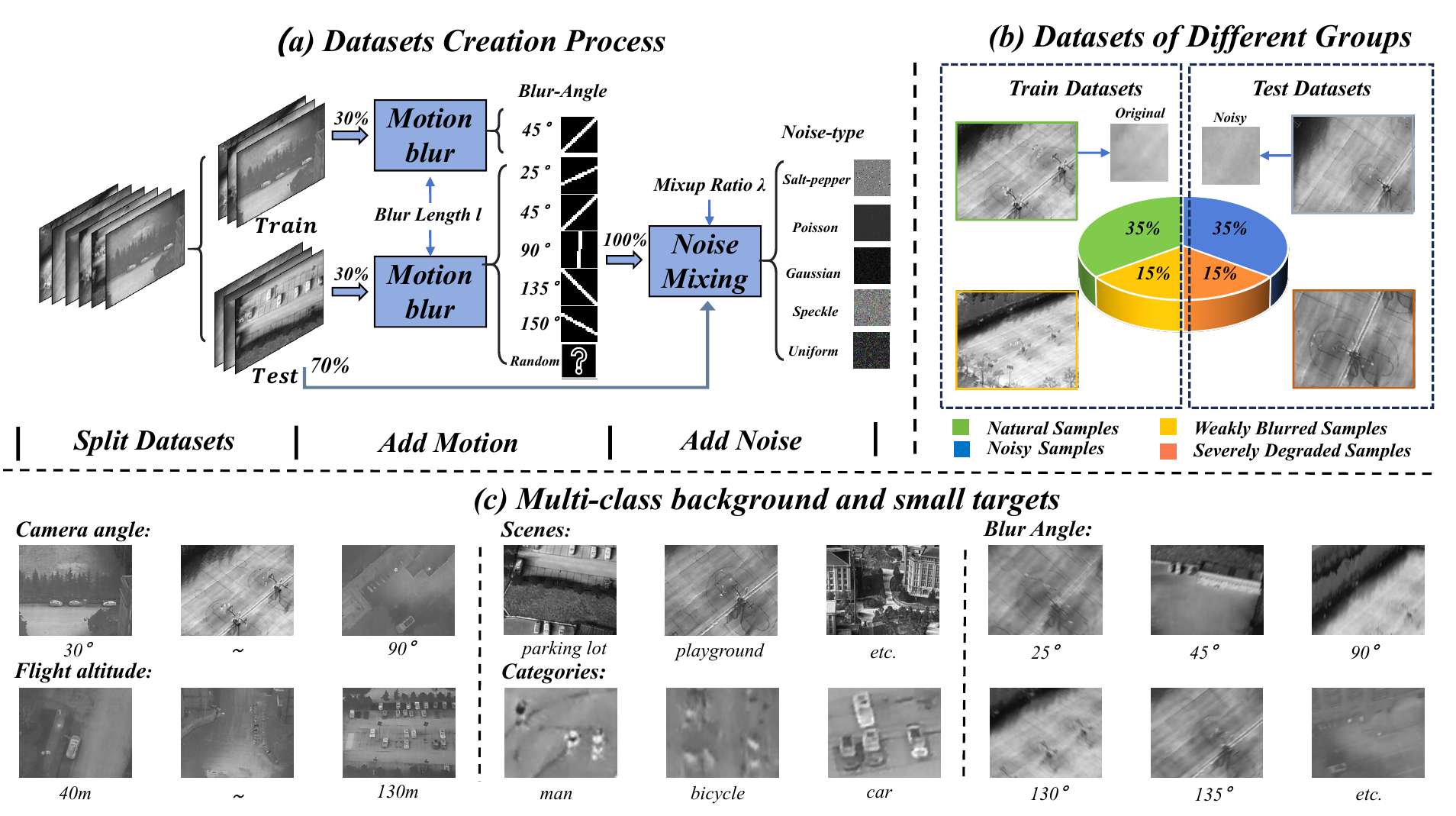}
\caption{(a) The diagram illustrates the process of constructing the dataset. (b) The diagram displays the composition and distribution of the dataset, where Severely Degraded Samples are created by combining Strongly Blurred Samples with Noisy Samples. (c) The diagram provides examples of backgrounds and targets across various scenes.}
\label{fig3}
\vspace{-3mm}
\end{figure*}

\textbf{Real-World Noise Guided Regularization.} To simulate real noise distributions, we build a noise mixing model. First, $k$ groups of images with consistent noise distributions are selected from the real target domain to create the noise sample library $X^{target}_k=[x^{target}_1,x^{target}_2,\dots,x^{target}_k]\in \mathbb R^{k\times c\times h\times w}$. Using the sliding window, we divide these $k$ groups of noise samples $X^{target}_k$ into multiple sampling regions $A_k$. Then, based on the local variance threshold $\sigma_{max}$ and mean threshold $\mu_{\text{min}}$, we adaptively select the noise regions $A^{
noise}_k=[a^{noise}_1,a^{noise}_2,\dots,a^{noise}_k]\in \mathbb R^{k\times c\times \frac{h}{15}\times \frac{w}{12}}$. Finally, these selected noise regions $A^{noise}_{k}$ are upsampled to the original resolution, creating the noise library $P^{noise}_k\in \mathbb{R}^{k\times c\times h\times w}$. At the same time, mixed samples $X^{noise}$ are generated using linear interpolation.
\begin{equation}
{X^{noise}} = \lambda \cdot \text {R}(P^{noise}_k) + (1 - \lambda) X^{source}  ,  \quad  \lambda \sim U(0,1).
\end{equation}
Here, $R(\cdot)$ refers to the random sampling from the noise library. The parameter $\lambda$ controls the noise injection intensity, and through our experiments, $\lambda=0.5$ was found to provide the best performance.

The final output layer $D_{1}\in \mathbb{R}^{H \times W \times 1}$of the encoder-decoder architecture is selected as the feature alignment node. A dual-branch network with shared weights is designed: the main branch processes the original input $X^{source}$ to extract clean features $\hat{Y}$, while the auxiliary branch processes noisy samples ${X^{noise}}$ to capture noise features $\tilde{Y}$. To enforce feature distribution consistency, a feature-level self-supervised loss function is defined.
\begin{equation}
\mathcal{L}_{\text SUP} = {{L}_{\text MSE}( \hat {Y}, \tilde{Y})}.
\end{equation}
Here,we set $\epsilon = 10^{-8}$ for numerical stability. $N$ denotes the batch size. This enables implicit domain adaptation to the test-time noise. We then optimize total loss function as:
\begin{equation}
\mathcal{L}_{\text{total}} = {L}_{\text{BCE}}
 + {L}_{\text{SUP}}.
\end{equation}

\begin{table*}[t]
\centering
\caption{Quantitative results of different models trained on NUAA and tested on IRSTD-1K. The optimal, second-optimal, and third-optimal values are labelled in red, blue, and green respectively.}
\scriptsize
\resizebox{\textwidth}{!}{
\begin{tabular}{lcccccc}
\hline
\multirow{3}{*}{Model} & \multicolumn{6}{c}{Train on NUAA} \\ \cline{2-7} 
                    & \multicolumn{6}{c}{Test on IRSTD-1K }                                     \\ \cline{2-7}  
                     & PixAcc$\uparrow$ ($\times 10^{-2}$) & mIoU$\uparrow$ ($\times 10^{-2}$) & nIoU$\uparrow$ ($\times 10^{-2}$) & $P_d\uparrow$ ($\times 10^{-2}$) & $F_a\downarrow$ ($\times 10^{-6}$) & F1$\uparrow$ ($\times 10^{-2}$) \\ \hline
ACM-Net~\cite{ACM} & 61.03 & \textcolor{blue}{50.87} & 51.12 & 88.92 & \textcolor{red}{30.68} & \textcolor{green}{55.49} \\
ALC-Net~\cite{ALC} & \textcolor{red}{89.16} & 25.02 & 26.19 & \textcolor{green}{90.60} & 519.76 & 39.07 \\
DNA-Net~\cite{DNA} & 53.80 & 46.85 & \textcolor{blue}{57.49} & 84.22 & \textcolor{blue}{30.89} & 50.08 \\
RDIAN~\cite{RDIAN} & 51.34 & 27.32 & 47.87 & 86.91 & 247.10 & 35.66 \\
ISTDU-Net~\cite{ISTDU-Net} & 63.10 & 47.18 & 53.00 & 87.91 & 86.71 & 53.99 \\
UIU-Net~\cite{UIU} & 60.24 & 48.30 & \textcolor{green}{56.82} & 87.91 & \textcolor{green}{50.54} & 53.61 \\
HCF-Net~\cite{hcf} & \textcolor{blue}{77.62} & 14.64 & 39.03 & \textcolor{blue}{91.89} & 963.54 & 24.63 \\
SCTransNet~\cite{sctransnet} & 60.16 & \textcolor{green}{49.15} & 55.66 & \textcolor{green}{90.60} & 60.77 & \textcolor{blue}{65.87} \\ \hline\hline
\textbf{Ours} & \textbf{\textcolor{green}{65.06}} & \textbf{\textcolor{red}{52.73}} & \textbf{\textcolor{red}{58.43}} & \textbf{\textcolor{red}{92.95}} & \textbf{55.49} & \textbf{\textcolor{red}{69.00}} \\
\hline
\end{tabular}
}
\label{tab1}
\end{table*}

\begin{table*}[t]
\centering
\caption{Quantitative results of different models trained on IRSTD-1K and tested on NUAA. The optimal, second-optimal, and third-optimal values are labelled in red, blue, and green respectively.}
\scriptsize
\resizebox{\textwidth}{!}{
\begin{tabular}{lcccccc}
\hline
\multirow{3}{*}{Model} & \multicolumn{6}{c}{Train on IRSTD-1K} \\ \cline{2-7} 
                    & \multicolumn{6}{c}{Test on NUAA }                                     \\ \cline{2-7}  
                     & PixAcc$\uparrow$ ($\times 10^{-2}$) & mIoU$\uparrow$ ($\times 10^{-2}$) & nIoU$\uparrow$ ($\times 10^{-2}$) & $P_d\uparrow$ ($\times 10^{-2}$) & $F_a\downarrow$ ($\times 10^{-6}$) & F1$\uparrow$ ($\times 10^{-2}$) \\ \hline
ACM-Net~\cite{ACM} & 83.22 & {66.30} & 65.23 & 92.15 & 96.07 & 36.38 \\
ALC-Net~\cite{ALC} & \textcolor{green}{85.84} & 60.00 & 65.20 & 94.11 & 102.85 & 70.63 \\
DNA-Net~\cite{DNA} & 83.10 & \textcolor{green}{73.59} & \textcolor{red}{77.24} & \textcolor{blue}{97.06} & \textcolor{red}{9.64} & 78.05 \\
RDIAN~\cite{RDIAN} & 76.93 & 57.56 & 68.37 & 91.18 & 115.13 & 73.06 \\
ISTDU-Net~\cite{ISTDU-Net} & 83.78 & 69.86 & 73.88 & \textcolor{green}{96.08} & 19.72 & 29.37 \\
UIU-Net~\cite{UIU} & \textcolor{blue}{86.21} & \textcolor{blue}{75.41} & \textcolor{green}{75.04} & 93.14 & \textcolor{blue}{11.10} & \textcolor{blue}{80.45} \\
HCF-Net\cite{hcf} & 63.29 & 53.30 & 54.15 & 87.01 & 31.79 & 57.82 \\
SCTransNet~\cite{sctransnet} & 79.92 & 65.96 & \textcolor{green}{75.04} & 95.10 & 61.66 & \textcolor{green}{79.48} \\ \hline\hline
\textbf{Ours} & \textbf{\textcolor{red}{87.45}} & \textbf{\textcolor{red}{75.44}} & \textbf{\textcolor{blue}{75.89}} & \textbf{\textcolor{red}{98.09}} & \textbf{\textcolor{green}{11.54}} & \textbf{\textcolor{red}{86.00}} \\
\hline
\end{tabular}
}
\label{tab2}
\end{table*}

\begin{table*}[t]
\centering
\caption{Comparison of target detection performance. This table presents the performance of both the state-of-the-art methods and our method across five evaluation metrics: pixel accuracy (PixAcc), mean intersection over union (mIoU), normalized intersection over union (nIoU), detection probability (Pd), and F1 score. The best results are marked in red.}
\scriptsize
\resizebox{\textwidth}{!}{
\begin{tabular}{lccccc}
\hline
\multirow{3}{*}{Method} 
                    & \multicolumn{5}{c}{Dynamic-ISTD Benchmark }                                     \\ \cline{2-6}  
                     & PixAcc$\uparrow$ ($\times 10^{-2}$) & mIoU$\uparrow$ ($\times 10^{-2}$) & nIoU$\uparrow$ ($\times 10^{-2}$) & Pd$\uparrow$ ($\times 10^{-2}$) & F1$\uparrow$ ($\times 10^{-2}$) \\ \hline
ACM-Net~\cite{ACM}                 & 19.46               & 17.62              & 19.52               & 27.34 & 18.49 \\
ALCNet~\cite{ALC}                 & 74.67               & 10.49              & 22.19                & 15.95 & 18.40 \\
DNA-Net~\cite{DNA}                 & \textcolor{blue}{78.41}               & \textcolor{green}{71.97}              & \textcolor{green}{56.46}                 & \textcolor{green}{56.71} & \textcolor{green}{75.05} \\
RDIAN~\cite{RDIAN}                   & 65.01               & 58.03              & 46.14                & 50.25 & 61.32 \\
ISTDU-Net~\cite{ISTDU-Net}               & 74.14  & 69.03 & 53.84                & 52.91 & 71.49 \\
UIU-Net~\cite{UIU}                 & 58.13              & 55.96              & 41.22              & 37.47 & 57.02 \\
HCF-Net\cite{hcf}              & 69.27              & 48.66              & 50.63              & 55.33 & 57.17 \\
SCTransNet~\cite{sctransnet}              & \textcolor{green}{78.37}  & \textcolor{blue}{74.51} & \textcolor{blue}{58.75}                & \textcolor{blue}{60.51} & \textcolor{blue}{84.47} \\ \hline \hline
\textbf{Ours}                     & \textbf{\textcolor{red}{81.70}} & \textbf{\textcolor{red}{76.01}}                                 & \textbf{\textcolor{red}{62.97}}                & \textbf{\textcolor{red}{68.10}} & \textbf{\textcolor{red}{85.51}} \\ \hline
    \end{tabular}
}
    \label{tab3}
\end{table*}

\section{Dynamic-ISTD Benchmark}
To address the challenges of real-world degradations, we introduces a cross-domain dynamic degradation dataset for drone infrared (Dynamic-ISTD Benchmark) in Fig.~\ref{fig3}. The dataset progressively degrades from the training set (source domain) to the test set (target domain) to simulate cross-domain shifts within the same dataset. It consists of 206 representative images selected from infrared aerial photography, we make pixel-level annotations to mark small targets. The image resolution is standardized at 640 $\times$ 512 and includes multiple scenes and three types of targets, captured under various conditions.We also proposes a dynamic degradation injection strategy to simulate drone flight observation conditions. Under this strategy, 70\% of the original data is retained as natural samples, while the remaining 30\% is processed with a motion blur degradation~\cite{yue2020supervised} to simulate extreme flight scenes.

For the training set $X_{train}$ the degradation parameters are set as follows: blur length $l=5$ and blur angle $\theta=45^\circ $ . To enhance the simulation of cross-domain shifts, the degradation method for the test set is more diverse. In the test set $X_{test}$, motion blur with discrete angles $\theta=[25^\circ,45^\circ,90^\circ,135^\circ,150^\circ,...]$ is applied to simulate the varying blur characteristics under different flight conditions. To further simulate the challenges of out-of-distribution noise, five types of noise, including Gaussian and salt-and-pepper noise, are combined and added to all images in the test set $X_{test}$ at a proportion of 0.05. This approach helps to evaluate the adaptability of the approach under cross-domain shift conditions more thoroughly.

\section{Experimental Validation}
In this section, we present our experimental dataset, training details, and comparative evaluations. We demonstrate the superiority of our proposed method through quantitative and qualitative assessments, along with ablation experiments to highlight the effectiveness of each module.

\subsection{Experiment setup}
\textbf{Implementation Details.} The experiments are implemented with the PyTorch, and all training and inference are conducted on an NVIDIA TitanXp GPU. To prevent the vanishing gradient problem, the network parameters are initialized using the Kaiming~\cite{he2015delving} normal distribution strategy. During training, the batch size is 8, and the Adam~\cite{adam} optimizer is used for parameter updates. The initial learning rate is 0.001 and is dynamically decayed to $10^{-5}$ using cosine annealing. The model is trained for 1000 epochs.

\textbf{Cross-dataset Settings.} To evaluate the domain robustness of the proposed method, we applied a strict cross-domain validation framework on our self-constructed Dynamic-ISTD benchmark. The dataset is evenly split into training (50\%) and test sets (50\%), with controlled domain shifts. Specifically, the training set introduces motion blur within a limited parameter space, while the test set contains multi-angle motion blur and composite noise perturbations. This setup significantly increases the difference between the training and test distributions, mimicking real-world domain variations. Additionally, to validate the generality and effectiveness of our cross-domain method, we performed bidirectional transfer experiments on two established benchmarks (NUAA-SIRST~\cite{NUAA-SIRST} and IRSTD-1k~\cite{IRSTD-1K}). All comparison methods were retrained using the same implementation details for a fair evaluation.

\subsection{Comparisons under the o.o.d condition}
Table.~\ref{tab1} and Table.~\ref{tab2} present the results of the cross-domain validation experiments. As shown, in the NUAA-SIRST~\cite{NUAA-SIRST} → IRSTD-1k~\cite{IRSTD-1K} cross-domain scenario, our method achieved a PixAcc of 65.06 and an F1-score of 69.00. Similarly, in the IRSTD-1k~\cite{IRSTD-1K} → NUAA-SIRST~\cite{NUAA-SIRST} cross-domain scenario, our method performed exceptionally well, with an F1-score of 86, outperforming all comparison methods. These cross-domain validation experiments clearly demonstrate that our method has strong generality and cross-domain adaptation capabilities. It can effectively handle the characteristic differences of different datasets, offering potential for cross-scenario deployment in real-world applications.

\subsection{Comparisons under the i.i.d condition}
\textbf{Quantitative Analysis.} To thoroughly assess the effectiveness of our proposed method, we compared it with several advanced methods in the infrared small target detection field, including ACM-Net~\cite{ACM}, ALC-Net~\cite{ALC}, DNA-Net~\cite{DNA}, RDIAN~\cite{RDIAN}, ISTDU-Net~\cite{ISTDU-Net}, UIU-Net~\cite{UIU}, HCF-Net\cite{hcf}, and SCTransNet~\cite{sctransnet}. As shown in Table.~\ref{tab3}, our method outperforms the existing methods in mIoU (76.01), nIoU (62.97), Pd (68.10), and F1 (85.51) scores.

\begin{figure*}[!t]
\centering
\includegraphics[width=0.95\linewidth]{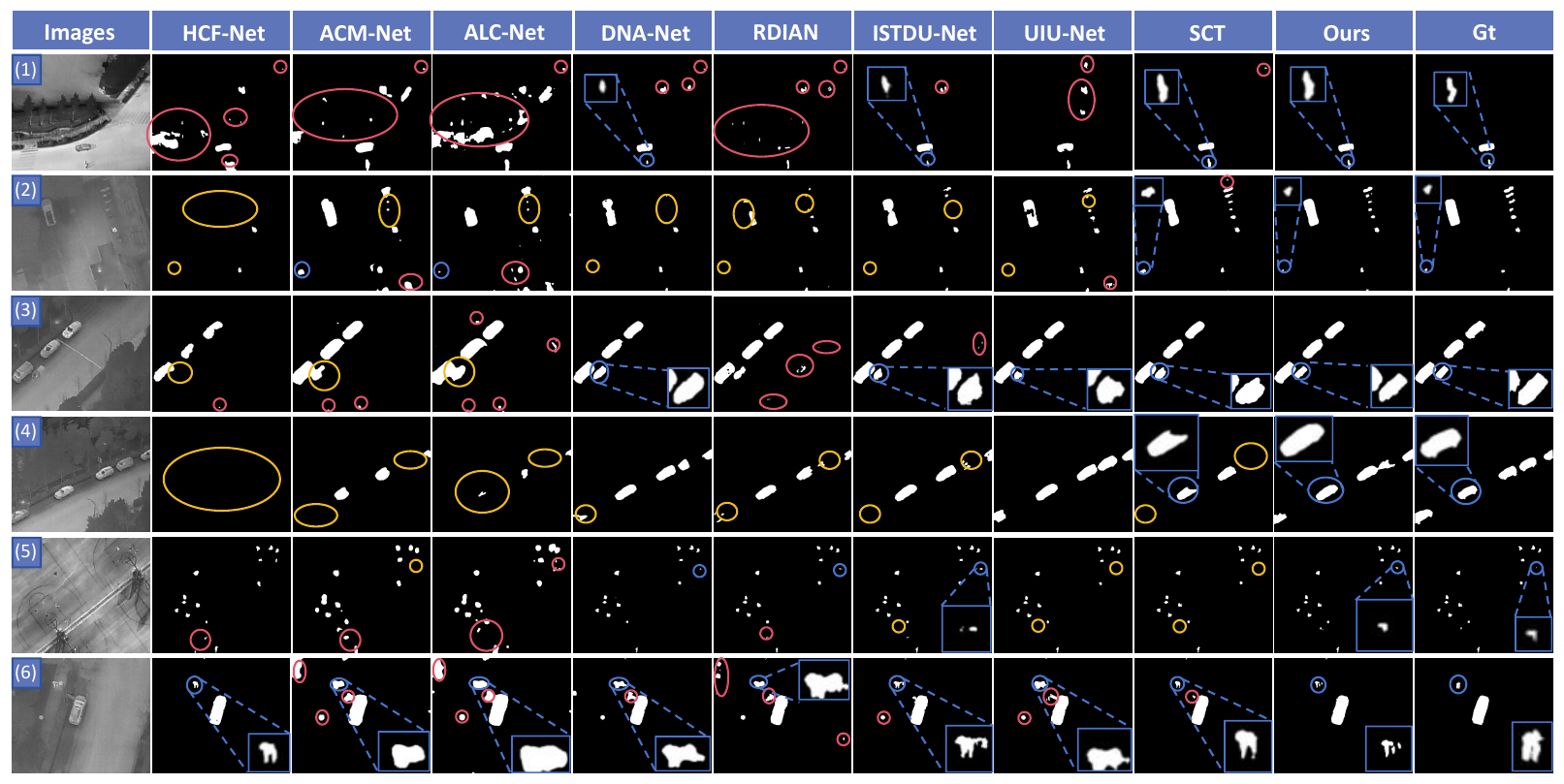}
\caption{Visualization of different models on the dataset. Blue, yellow, and red circles indicate correctly detected targets, missed detections, and false positives, respectively. The blue boxes highlight small targets for clearer observation.}
\label{fig9}
\vspace{-3mm}
\end{figure*}

\textbf{Visualization Analysis.} Fig.~\ref{fig9} shows the visual results of 8 representative algorithms on Dynamic-ISTD Benchmark. Our method significantly improves background region discrimination accuracy and achieves high-precision target localization, effectively solving the problem of distinguishing adjacent targets. In Fig.~\ref{fig9} (1), while the SCTransNet~\cite{sctransnet}  performs similarly to our method in target clarity, it has a much higher background misclassification rate, with other comparison models showing similar issues. In Fig.~\ref{fig9} (3), our method provides more accurate and clearer recognition of small car targets, matching real small targets more closely and avoiding the unclear contours seen in other models.

\textbf{Efficiency Analysis.} As shown in Table.~\ref{tab4}, our method strikes a much better balance between accuracy and efficiency than other models. Compared to SCTransNet~\cite{sctransnet} , which has a similar parameter count, our method increases IoU to 74.51 (+1.5) and reduces inference time by 0.66. In comparison to the efficient DNA-Net~\cite{DNA}, we achieve a 4.04 increase in IoU with slightly more parameters and longer inference time. Against UIU-Net~\cite{UIU}, with 50.54M parameters, our method delivers higher IoU with just 22\% of the parameters and twice the speed.

\begin{table}[t]
    \caption{The performance of different methods in terms of parameters, inference time, and IoU on the Dynamic-ISTD Benchmark.}
    \label{table4}
    \centering
    \scriptsize
    \resizebox{1\columnwidth}{!}
    {
    \begin{tabular}{lcccc}
    \hline
    \multirow{2}{*}{Method} & \multicolumn{4}{c}{Dynamic-ISTD Benchmark} \\ \cline{2-5} 
                         & Pub year                                 & Params (M)                                   & Inference times ($10^{-3}$s)                & IoU$\uparrow$ \\ \hline
    ACM-Net~\cite{ACM}                 & 2021               & 0.39              & 1.77               & 17.62 \\
    ALC-Net~\cite{ALC}                 & 2021               & 0.43              & 1.66                & 10.49 \\
    DNA-Net~\cite{DNA}                 & 2022               & 4.70              & 4.11                 & 71.97 \\
    RDIAN~\cite{RDIAN}                   & 2022               & 2.75              & 1.35                & 58.03 \\
    ISTDU-Net~\cite{ISTDU-Net}               & 2022  & 2.75 & 3.52                & 69.03 \\
    UIU-Net~\cite{UIU}                 & 2022              & 50.54              & 4.07              & 55.96 \\
    HCF-Net\cite{hcf}              & 2024              & 0.22              & 8.35              & 48.66 \\
    SCTransNet~\cite{sctransnet}              & 2024  & 11.19 & 5.57                & 74.51 \\ \hline  \hline
    \textbf{Ours}                    & \textbf{2025} & \textbf{11.33 }                                &\textbf{ 6.23}                & \textbf{{\color{red}{ \textbf{76.01}}}} \\ \hline
    \end{tabular}
    }
    \label{tab4}
\end{table}

\begin{table}[ht]
    \caption{Ablation study on different Background Region Detection Methods on the Dynamic-ISTD Benchmark.}
    \label{table5}
    \centering
    \scriptsize
    \resizebox{1\columnwidth}{!}
    {
    \begin{tabular}{lccccc}
    \hline
    \multirow{2}{*}{Method} & \multicolumn{5}{c}{Dynamic-ISTD Benchmark} \\ \cline{2-6} 
                         & PixAcc$\uparrow$                                 & mIoU$\uparrow$                                   & nIoU$\uparrow$                & Pd$\uparrow$ & F1$\uparrow$ \\ \hline
    Baseline                 & 78.37               & 74.51              & 58.75               & 60.51 & 84.47 \\
    Random Crop                 & 77.78               & 73.98              & 59.58                & 65.32 & 84.12 \\
    w/o WMF                 & 79.51               & 74.78              & 60.10                 & 68.61 & 84.17 \\
    \textbf{Ours}                   & \textbf{{\color{red}{80.37}} }              & \textbf{{\color{red}{75.26}} }             & \textbf{{\color{red}{62.23}} }               & \textbf{{\color{red}{68.23}}} & \textbf{{\color{red}{84.98}}} \\ \hline
    \end{tabular}
    }
    \label{tab5}
\end{table}

\begin{table}[ht]
    \caption{Ablation study on varying the hyperparameter $\alpha$ within the Real-world Domain Noise Invariance Learning framework.}
    \label{table6}
    \centering
    \scriptsize
    \resizebox{1\columnwidth}{!}
    {
    \begin{tabular}{lccccc}
    \hline
    \multirow{2}{*}{Method} & \multicolumn{5}{c}{Dynamic-ISTD Benchmark} \\ \cline{2-6} 
                         & PixAcc$\uparrow$                                 & mIoU$\uparrow$                                   & nIoU$\uparrow$                & Pd$\uparrow$ & F1$\uparrow$ \\ \hline
    $\alpha$ = 0.1                  & 78.90               & 74.48              & 60.38               & 63.80 & 84.49 \\
    $\alpha$ = 0.3                & 79.90               & 74.68              & 60.53                & 67.85 & 84.63 \\
    $\alpha$ = 0.5                 & {\color{red}{81.70}}               & {\color{red}{76.01}}              & {\color{red}{62.94}}                 & {\color{red}{68.10}} & {\color{red}{85.51}} \\
    $\alpha$ = 0.7                   & 79.62               & 75.19              & 60.40                & 64.68 & 84.96 \\
    $\alpha$ = 0.9                   & 78.65               & 73.53              & 60.01                & 67.34 & 83.83 \\ \hline
    \end{tabular}
    }
    \label{tab6}
\end{table}

\begin{table}[ht]
    \caption{Noise Mixing Types on the Dynamic-ISTD Benchmark.}
    \label{table7}
    \centering
    \scriptsize
    \resizebox{1\columnwidth}{!}
    {
    \begin{tabular}{lccccc}
    \hline
    \multirow{2}{*}{Noise-Type} & \multicolumn{5}{c}{Dynamic-ISTD Benchmark} \\ \cline{2-6} 
                         & PixAcc$\uparrow$                                 & mIoU$\uparrow$                                   & nIoU$\uparrow$                & Pd$\uparrow$ & F1$\uparrow$ \\ \hline
    Baseline                 & 78.37               & 74.51              & 58.75               & 60.51 & 84.47 \\
    Gaussian                 & 74.84               & 72.04              & 55.14                & 58.61 & 82.70 \\
    Salt-Pepper                 & 77.96               & 74.71              & 59.89                 & 64.56 & 84.58 \\
    Speckle                   & 78.36               & 75.21              & 59.58                & 63.42 & 84.88 \\
    Uniform                   & 78.02               & 74.15              & 59.06                & 61.51 & 84.22 \\
    Poisson                   & 79.30               & 75.37              & 59.55                & 64.94 & 85.03 \\
    Composite Noise                   & 79.52               & 74.92              & 61.10                & 66.08 & 84.79 \\
    Real-World Noise                   & {\color{red}{79.75}}               & {\color{red}{75.63}}              & {\color{red}{60.81}}                & {\color{red}{67.72}} & {\color{red}{85.22}} \\ \hline
    \end{tabular}
    }
    \label{tab7}
\end{table}

\begin{table}[t]
    \caption{Ablation study of Fusion datasets and Self-supervised modules on the Dynamic-ISTD Benchmark.}
    \label{table8}
    \setlength{\tabcolsep}{1pt}
    \centering
    \scriptsize
    \resizebox{1\columnwidth}{!}
    {
    \renewcommand{\arraystretch}{1.5}
    \begin{tabular}{cccccccc}
    \hline
    \multicolumn{1}{l}{\multirow{2}{*}{Baseline}} & \multirow{2}{*}{Fusion datasets} & \multirow{2}{*}{Self-supervised} & \multicolumn{5}{c}{Performance}             \\ \cline{4-8} 
    \multicolumn{1}{l}{}                          &                      &                        & PixAcc$\uparrow$         & mIoU$\uparrow$          & nIoU$\uparrow$           & Pd$\uparrow$ & F1$\uparrow$ \\ \hline
    $\checkmark$                                             & $\times$                     & $\times$                      & 78.37        & 74.51        & 58.75         & 60.51 & 84.47 \\
    $\checkmark$                                             & $\checkmark$                    & $\times$                     & 80.37        & 75.26        & 62.23         & 68.23 & 84.98 \\
    $\checkmark$                                             & $\checkmark$                    & $\checkmark$                                     & \textbf{\textcolor{red}{81.70}}\color{green}{(+3.33)}        & \textbf{\textcolor{red}{76.01}}\color{green}{(+1.50)}        & \textbf{\textcolor{red}{62.97}}\color{green}{(+4.22)}         & \textbf{\textcolor{red}{68.10}}\color{green}{(+7.59)} & \textbf{\textcolor{red}{85.51}}\color{green}{(+1.04)} \\ \hline
    \end{tabular}
    }
    \label{tab8}
\end{table}

\subsection{Ablation Studies}

\textbf{Effectiveness of Background Region Detection.} The effectiveness of Background Region Detection comes from the combination of wavelet filtering and probabilistic clipping. These reduce random noise interference in segmentation while preserving key structural information. Wavelet filtering improves feature extraction by separating low-frequency basebands from high-frequency details, and probabilistic clipping precisely locates the key regions. As shown in Table~\ref{tab5}, Background Region Detection enhances feature extraction and significantly improves segmentation accuracy.

\textbf{Self-supervised Noise Consistency Parameters.} Table~\ref{tab6} shows an important balance point in the noise-guided adaptation mechanism. The best performance is achieved when $\alpha= 0.5$, indicating that the model needs to retain enough original image information while introducing sufficient noise variation in the learning process. Lower values of$\alpha$(e.g., 0.1, 0.3) may not introduce enough noise for the model to learn adaptive features, while higher values (e.g., 0.7, 0.9) could add too much noise, disrupting the key structures of the original image. At $\alpha= 0.5$, the model effectively combines features consistent with the original image, while also learning noise robustness, achieving the best overall performance.

\textbf{Noise Tpye.} Table~\ref{tab7} shows the impact of different noise types on model performance. Composite Noise refers to a complex noise formed by combining five types of noise, including Gaussian noise, Salt-Pepper noise and so on. It performs well in nIoU, suggesting that this noise type aids in improving segmentation accuracy and target localization. Real-World Noise delivers the best performance, with a PixAcc of 79.75 and an F1 score of 85.22, highlighting that the model with added noise has the highest robustness.

\textbf{Ablation on Fusion datasets and Self-supervised.} As shown in Table~\ref{tab8}, the Baseline performance is significantly lower than that of the other experimental groups. Re-Generation diversifies small target scenarios and greatly boosts PixAcc. The noise consistency self-supervised strategy enhances the model's robustness to noise, effectively reducing the negative impact of noise in the test set. Overall, the collaborative optimization of multiple strategies helps overcome the performance limitations of individual modules.

\section{Conclusion}
We present a domain adaptation enhancement framework to tackle the cross-domain distribution shift problem in infrared small target detection. Wavelet-guided Cross-domain Synthesis improves adaptability to target environments without requiring extra inference adjustments. Real-world Domain Noise Invariance Learning overcomes the limitations of artificial noise assumptions and boosts robustness to heterogeneous noise. Cross-domain validation experiments conducted on both custom and real datasets show the effectiveness and wide applicability of our method across different datasets.

\bibliographystyle{IEEEtran}
\bibliography{ref}

\begin{IEEEbiography}[{\includegraphics[width=1in,height=1.25in,clip,keepaspectratio]{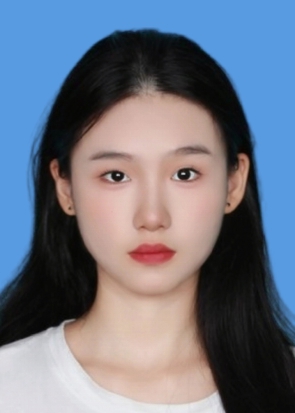}}]{Yuehui Li} is currently pursuing the B.S. degree at the School of Information Engineering, Guangdong University of Technology, Guangzhou, China. Her research interests include computer vision and machine learning.
\end{IEEEbiography}

\begin{IEEEbiography}[{\includegraphics[width=1in,height=1.25in,clip,keepaspectratio]{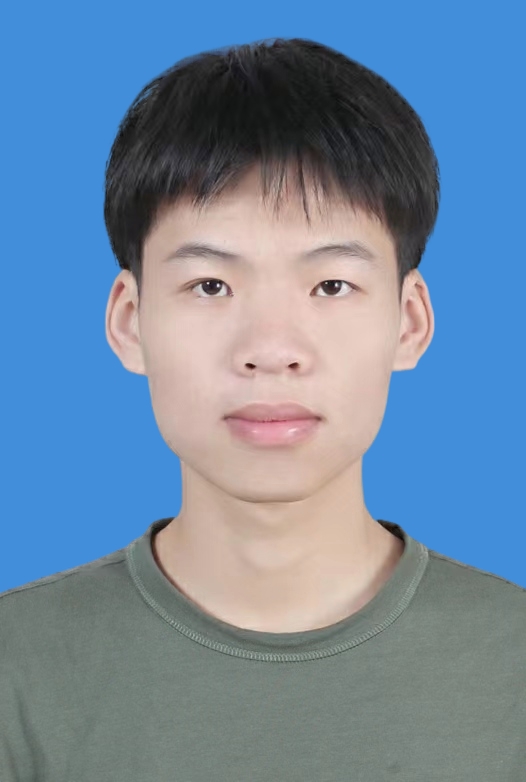}}]{Yahao Lu} received the B.S. degree in 2023, from the School of Information Engineering, Guangdong University of Technology, Guangzhou, China, where he is currently working towards a M.S. degree. His research interests include computer vision and machine learning.
\end{IEEEbiography}

\begin{IEEEbiography}[{\includegraphics[width=1in,height=1.25in,clip,keepaspectratio]{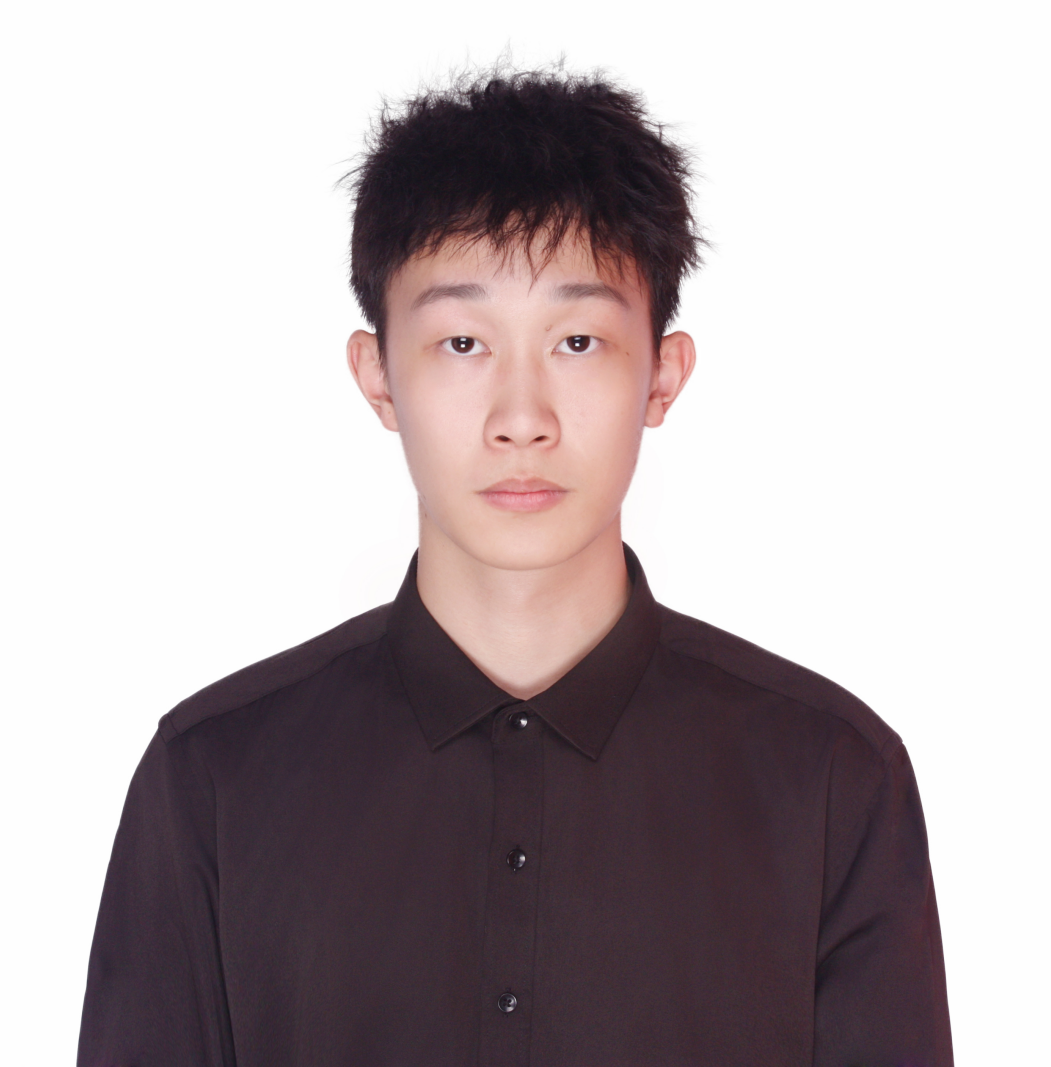}}]{Haoyuan Wu} is currently pursuing a B.S. in Communication Engineering at Guangdong University of Technology, Guangzhou, China. His academic interests are centered in robotics engineering and machine vision.
\end{IEEEbiography}

\begin{IEEEbiography}
[{\includegraphics[width=1in,height=1.25in,clip,keepaspectratio]{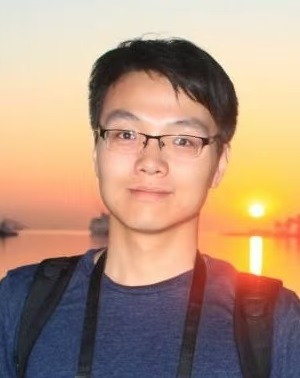}}]{Sen Zhang}
received the Ph.D. degree from the School of Computer Science, the University of Sydney, Australia, in 2023. Previously, he obtained a B.S. degree in 2014, from the School of Biomedical Engineering, Tsinghua University, China. He is currently a Researcher at ByteDance Inc, Australia. His research interests include reinforcement learning, computer vision and large language model.
\end{IEEEbiography}

\begin{IEEEbiography}
[{\includegraphics[width=1in,height=1.25in,clip,keepaspectratio]{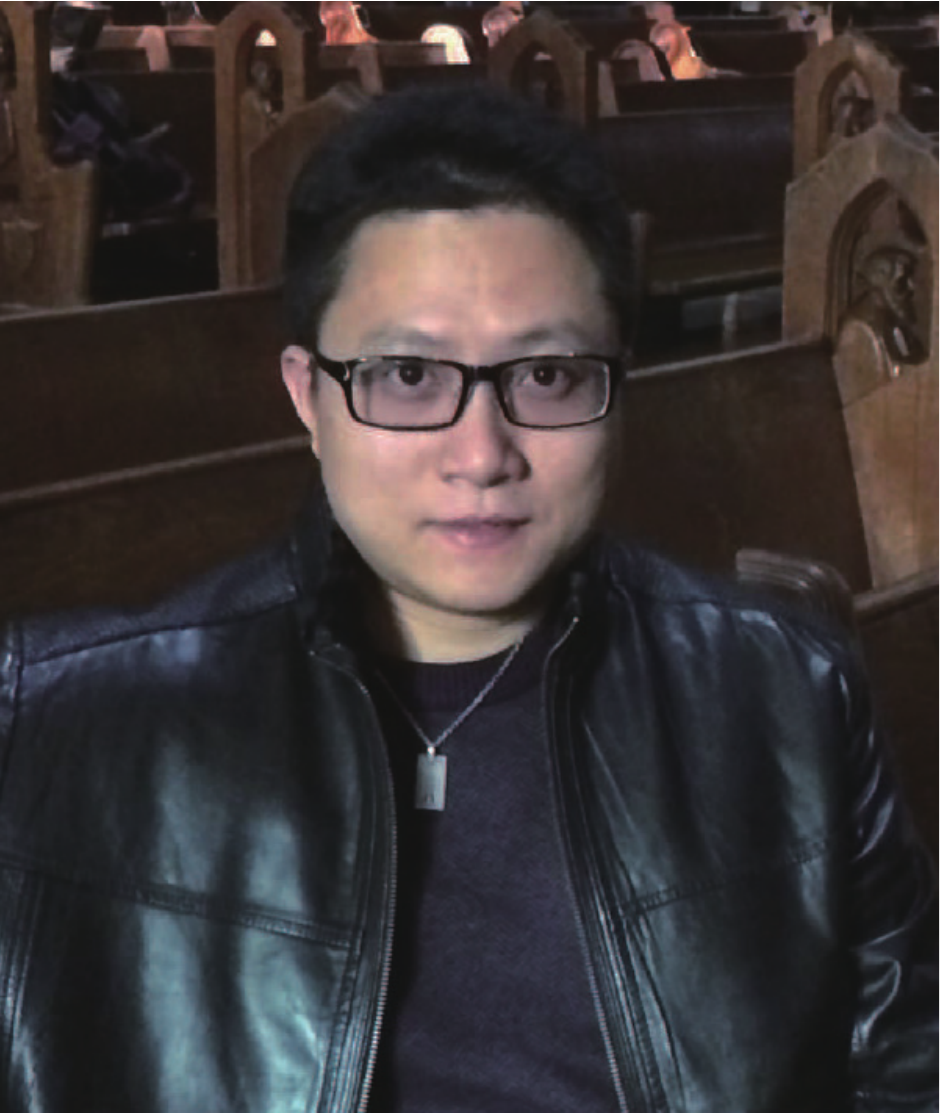}}]{Liang Lin}(Fellow, IEEE) is a Full Professor of computer science at Sun Yat-sen University. He served as the Executive Director and Distinguished Scientist of SenseTime Group from 2016 to 2018, leading the R$\&$D teams for cutting-edge technology transferring. He has authored or co-authored more than 200 papers in leading academic journals and conferences, and his papers have been cited by more than 26,000 times. He is an associate editor of IEEE Trans.Neural Networks and Learning Systems and IEEE Trans. Multimedia, and served as Area Chairs for numerous conferences such as CVPR, ICCV, SIGKDD and AAAI. He is the recipient of numerous awards and honors including Wu Wen-Jun Artificial Intelligence Award, the First Prize of China Society of Image and Graphics, ICCV Best Paper Nomination in 2019, Annual Best Paper Award by Pattern Recognition (Elsevier) in 2018, Best Paper Dimond Award in IEEE ICME 2017, Google Faculty Award in 2012. His supervised PhD students received ACM China Doctoral Dissertation Award, CCF Best Doctoral Dissertation and CAAI Best Doctoral Dissertation. He is a Fellow of IEEE/IAPR/IET.
\end{IEEEbiography}

\begin{IEEEbiography}
[{\includegraphics[width=1in,height=1.25in,clip,keepaspectratio]{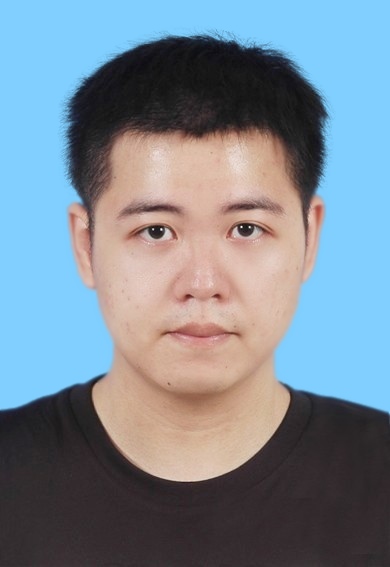}}]{Yukai Shi}
received the Ph.D. degrees from the school of Data and Computer Science, Sun Yat-sen University, Guangzhou China, in 2019. He is currently an associate professor at the School of Information Engineering, Guangdong University of Technology, China. His research interests include computer vision and machine learning.
\end{IEEEbiography}

\end{document}